\documentclass[11pt]{article}

\usepackage{acl2020}
\usepackage{enumitem}
\setlist{nosep}

\usepackage{times}
\usepackage{amsmath}
\usepackage{multirow}
\usepackage{booktabs}
\usepackage{tikz}
\usepackage{CJK}
\usepackage{hhline}

\aclfinalcopy
\newcommand{\model}{INSET}

\usepackage{array}
\newcolumntype{H}{>{\setbox0=\hbox\bgroup}c<{\egroup}@{}}

\begin{document}

\begin{CJK*}{UTF8}{}

\title{INSET: Sentence Infilling with INter-SEntential Transformer}

\CJKfamily{gbsn}

\author{Yichen Huang (黄溢辰)\textsuperscript{12}\Thanks{~These authors contributed equally to this work.}~, Yizhe Zhang\textsuperscript{1}\footnotemark[1]~, Oussama Elachqar\textsuperscript{1}, Yu Cheng\textsuperscript{1}\\
  \textsuperscript{1}Microsoft Corporation, Redmond, Washington 98052, USA\\
  \textsuperscript{2}Center for Theoretical Physics, MIT, Cambridge, Massachusetts 02139, USA\\
  \texttt{yichuang@mit.edu, \{yizzhang, ouelachq, yu.cheng\}@microsoft.com}}

\date{}

\maketitle

\end{CJK*}

\begin{abstract}

Missing sentence generation (or sentence infilling) fosters a wide range of applications in natural language generation, such as document auto-completion and meeting note expansion. This task asks the model to generate intermediate missing sentences that can syntactically and semantically bridge the surrounding context. Solving the sentence infilling task requires techniques in natural language processing ranging from understanding to discourse-level planning to generation. In this paper, we propose a framework to decouple the challenge and address these three aspects respectively, leveraging the power of existing large-scale pre-trained models such as BERT and GPT-2. We empirically demonstrate the effectiveness of our model in learning a sentence representation for generation and further generating a missing sentence that fits the context.

\end{abstract}

\section{Introduction}

\begin{figure}[t!]
\centering
\includegraphics[width=\linewidth]{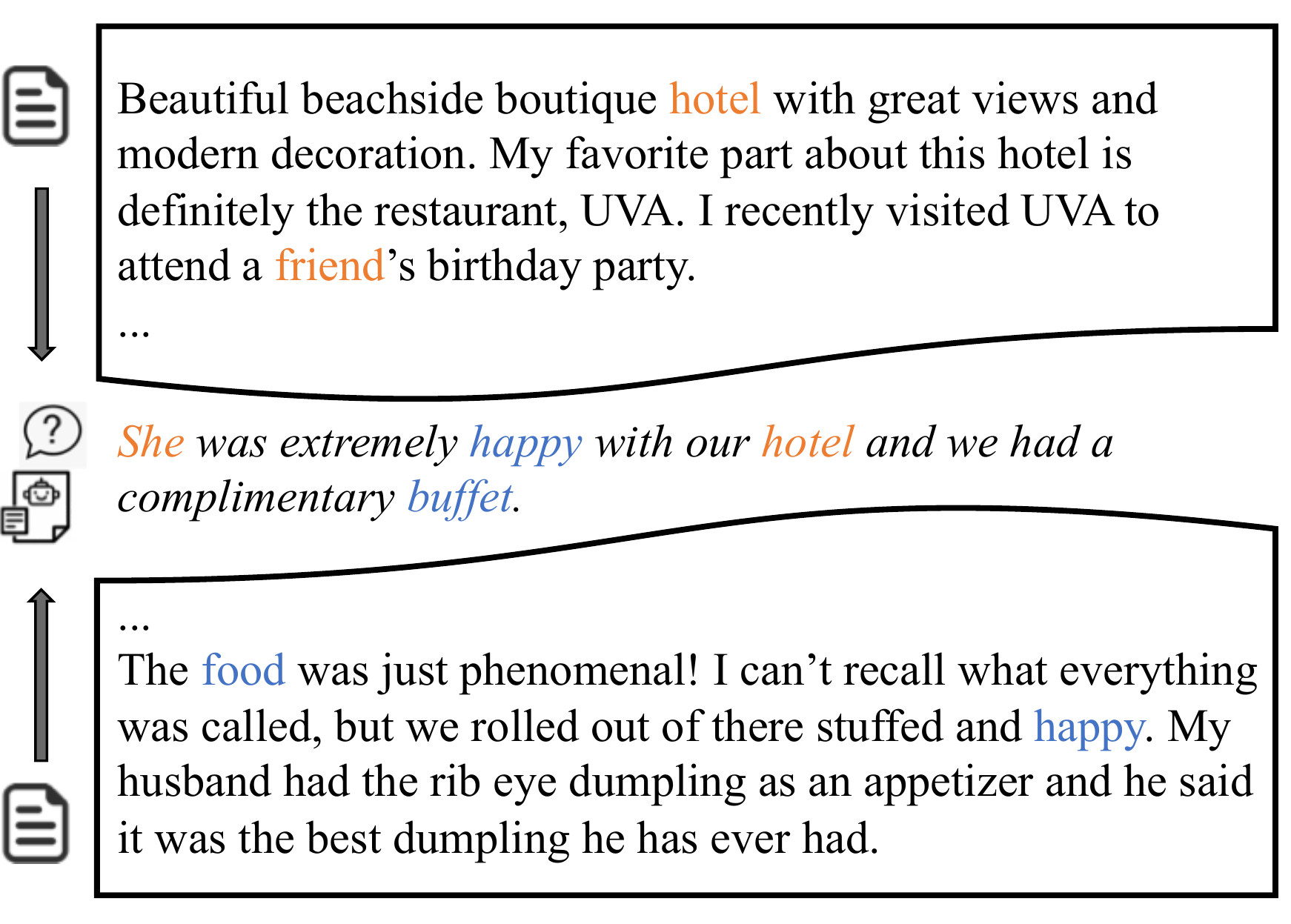}
\caption{Sentence infilling: generating an intermediate sentence that provides a smooth semantic transition from the preceding to the following context. This example is generated by our model on the TripAdvisor dataset. The colors mark the correspondence between the generated sentence and the context.}
\label{fig:task}
\end{figure}

Generating a span of missing tokens in a text chunk, known as ``text infilling,'' has attracted many attentions recently \cite{ZHX19, STQ+19, liu-etal-2019-tigs, TGCE19, JCL+19}. Here we study the related but somewhat different task of ``sentence infilling.'' Specifically, as illustrated in Figure~\ref{fig:task}, intermediate sentences (or chunks of text) are removed from long-form text (e.g., paragraphs, documents), and the task is to generate the missing pieces that can smoothly blend into and fit the context both \emph{syntactically} and \emph{semantically}. The generation can be either based only on context, or based on both context and side information such as constraint keywords. Compared with text infilling, sentence infilling requires the model to handle inter-sentential correlation and to reason about missing semantic information. Developing models for sentence infilling can potentially facilitate many text generation applications. Possible scenarios include, but are not limited to: document auto-completion by detecting and suggesting missing bridging sentences in the surrounding context; collaborative document writing by modifying and unifying different writing styles from multiple authors; meeting note expansion by extending a set of keywords (lexical constraints) to a full sentence, leveraging the surrounding context.

There are many challenges associated with this long-form sentence infilling task, which is typically a one-to-many problem in that the possible outputs can be diverse. As the generated sentence should connect separate text pieces in a syntactically and semantically smooth and coherent manner, the task requires a wide range of \emph{understanding}, \emph{planning}, and \emph{generation} techniques. Large-scale pre-trained language models such as BERT \cite{bert} and GPT-2 \cite{radford2019language} have dramatically enhanced the understanding and generation modules. However, how to integrate them in a holistic manner, and to analyze and establish the long-range dependence structure by high-level semantic planning is still challenging and yet to explore, as semantic appropriateness is usually subtler than syntactic appropriateness, which can be well characterized by autoregressive language models.

Several works have been done in this direction. MASS~\cite{STQ+19} obtains sentence representations by predicting a span of missing tokens. It can be used to generate missing text, but the missing span length needs to be pre-specified. Other related works \cite{liu-etal-2019-tigs, JCL+19} also require knowledge of the span length as an input to their models, and thus are different from our work. Text infilling~\cite{ZHX19} sequentially generates tokens for the missing part of a sentence until an end-of-blank token is generated. Its generation can be of arbitrary length. By design, all these previous approaches operate at the token level, and thus arguably focus more on lexical appropriateness than the global semantics.

In this paper, we propose INter-SEntential Transformer (\model), a novel approach to sentence infilling. Our model first produces sentence-level semantic features that capsulate the missing high-level information. Then, grounded on the predicted semantic features, the model generates the syntactic and lexical features to embody the predicted sentence. Specifically, \emph{understanding}, \emph{planning}, and \emph{generation} are handled by three modules in a synergistic manner:
\begin{itemize}
\item a \emph{BERT-based encoder} to map each sentence to the latent semantic space.
\item a \emph{sentence-level semantic planner} to infer the missing information that can bridge the semantics of preceding and following context.
\item a \emph{GPT-based generator} (decoder) to map semantic features back to the text domain.
\end{itemize}

The main contributions and advantages of this work are summarized as follows:
\begin{itemize}
\item We study the task of sentence infilling, which requires the model to handle inter-sentential correlation and to predict missing semantic information. This goes beyond text infilling \cite{ZHX19}, which asks the model to fill in the missing part of a single sentence.
\item Our approach decouples understanding, planning, generation, and leverages existing large-scale pre-trained understanding and generation models (BERT, GPT-2). The components of our model can be separately examined and improved with additional data.
\item Our model predicts a feature vector in the latent semantic space for the missing sentence and maps the vector to text. Thus, it takes care of semantic smoothness and appropriateness.
\item Our model allows the generation to be of arbitrary length, as in \cite{ZHX19}.
\item Compared with directly processing text, our approach significantly reduces computation time and memory usage during training, as (after pre-computing sentence features) the sequence length is the number of sentences rather than that of tokens.
\end{itemize}

\section{Related Work}

\paragraph{Pre-Trained Language Model.} Language models pre-trained on a large corpus improve natural language understanding and generation through transferable contextualized word representations \cite{bert, lample2019large} and models \cite{howard-ruder-2018-universal}. Large transformer models \cite{Vaswani2017} like GPT-2 \cite{radford2019language}, Megatron (\url{https://github.com/NVIDIA/Megatron-LM}), and T5 \cite{RSR+19} can achieve state-of-the-art results without training on any particular language modeling benchmark. \cite{keskarCTRL2019} proposes a conditional generation model, trained to condition on control codes that govern style, content, and other task-specific properties. Different from them, our model builds sentence representations via autoencoding with a pair of BERT and GPT-2.

\paragraph{Context-Aware Text Generation.} There are some related works on context-aware text generation \cite{Mikolov2012ContextDR, Tang2016ContextawareNL, mangrulkar-etal-2018-context}. Most previous works on language modeling with contextual information \cite{DBLP:journals/corr/WangC15g,wang2017topic, sordoni-etal-2015-neural,wen-etal-2015-stochastic, DBLP:journals/corr/VinyalsL15} treat the preceding sentences as context. Compared with these sequential generation tasks, our task is constrained by bidirectional context, and is more challenging.

Text infilling \cite{ZHX19} aims at filling in the missing part, given the rest of a sentence. \cite{liu-etal-2019-tigs} proposes an iterative inference algorithm based on gradient search for text infilling. For story infilling, \cite{TGCE19} first predicts rare words in the missing span, and then generates text conditioned on these words. SpanBERT \cite{JCL+19} masks random contiguous spans and (pre-)trains a language model to predict tokens in the span. XL-Editor \cite{SCY19} adapts XLNet \cite{YDY+19} to text infilling and other editing tasks.

\cite{kang2019linguistic} models logic connections between sentences and generates intermediate sentences grounded on inter-sentential ``flow.'' \cite{aakur2019abductive} formulates abductive commonsense reasoning as a natural language inference task to decide the appropriate reason that could explain the observation in one sentence given the background described by another sentence. \cite{cheng2020contextual} proposes a text style transfer task to translate a sentence in the context of a paragraph into the desired style. These three works study generation tasks that address inter-sentential relationship, and thus may be conceptually related to our motivation.

Compared with \cite{ZHX19, liu-etal-2019-tigs, TGCE19, JCL+19, SCY19, kang2019linguistic, aakur2019abductive, cheng2020contextual}, our approach is clearly different. We fully exploit existing large-scale pre-trained models BERT and GPT-2 to learn smooth sentence embeddings in the latent semantic space, and then process sentence-level information in this space.

\paragraph{Hierarchical Text Generation.} Hierarchical text generation with high-level semantic planning has been studied in many previous works. \cite{Sordoni:2015:HRE:2806416.2806493} presents a hierarchical recurrent encoder-decoder architecture for context-aware query suggestion. \cite{zhang2019consistent} proposes a framework to infer semantic features for response generation using self-supervised learning. Previous works have used multi-level LSTM encoders \cite{yang2016hierarchical,hu2019makes} and hierarchical autoencoders \cite{li2015hierarchical} to learn hierarchical representations for long text. \cite{shen2019towards} uses a variational autoencoder to encode an entire paragraph into a single latent variable, from which the paragraph can be generated hierarchically. In comparison, our task is to generate intermediate sentences in the surrounding context.

\section{Tasks and Methods} \label{ref:cast}

\subsection{Task Definition}  \label{sec:task_definition}

The task of sentence infilling is formally defined as follows. Consider a dataset of $N$ paragraphs $\{p^{(k)}\}_{k=1}^N$. Each paragraph $p^{(k)}=(s^{(k)}_1, s^{(k)}_2, \ldots, s^{(k)}_{M_k})$ consists of $M_k$ consecutive sentences. For each $k$, we are given a positive integer $m_k\le M_k$ and the context $(s^{(k)}_1, s^{(k)}_2, \ldots, s^{(k)}_{m_k-1}, s^{(k)}_{m_k+1}, \ldots, s^{(k)}_{M_k})$, but the $m_k$'th sentence $s^{(k)}_{m_k}$ is missing. The task is to generate a sentence $\hat s^{(k)}_{m_k}$ in the missing position such that it fits the context. For simplicity and without any confusion, we drop the index $k$ from now on (note that $M$ and $m$ may depend on $k$).

The criteria for successful generation are:
\begin{itemize}
\item The sentence $\hat s_m$ is fluent and meaningful.
\item Inserting the generated sentence into the context, we obtain a semantically coherent paragraph $(s_1, s_2, \ldots, s_{m-1}, \hat s_m, s_{m+1}, \ldots, s_M)$.
\item $\hat s_m$ is written in the same style as contextual sentences $\{s_j\}_{j\neq m}$.
\end{itemize}

Since there could be multiple semantically different sentences that fit the same context well, it is not necessary for $\hat s_m$ to be close to the ground truth $s_m$. Rather, $\hat s_m$ is considered successful as long as it satisfies the criteria above.

\subsection{\model: Inter-Sentential Transformer}

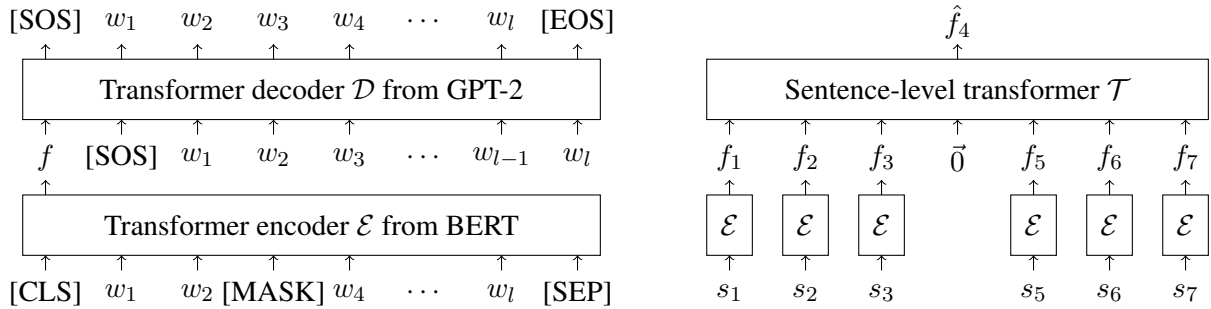
\begin{figure*}
    \centering
    \begin{tikzpicture}
    \node at (0, 0) {[CLS]};
    \node at (1, 0) {$w_1$};
    \node at (2, 0) {$w_2$};
    \node at (3, 0) {[MASK]};
    \node at (4, 0) {$w_4$};
    \node at (5, 0) {$\cdots$};
    \node at (6, 0) {$w_l$};
    \node at (7, 0) {[SEP]};
    \draw [->] (0, .3) -- (0, .5);
    \draw [->] (1, .3) -- (1, .5);
    \draw [->] (2, .3) -- (2, .5);
    \draw [->] (3, .3) -- (3, .5);
    \draw [->] (4, .3) -- (4, .5);
    \draw [->] (6, .3) -- (6, .5);
    \draw [->] (7, .3) -- (7, .5);
    \draw (-.3, 0.5) rectangle (7.3, 1.3);
    \node at (3.5, 0.9) {Transformer encoder $\mathcal E$ from BERT};
    \draw [->] (0, 1.3) -- (0, 1.5);
    \node at (0, 1.8) {$f$};
    \node at (1, 1.8) {[SOS]};
    \node at (2, 1.8) {$w_1$};
    \node at (3, 1.8) {$w_2$};
    \node at (4, 1.8) {$w_3$};
    \node at (5, 1.8) {$\cdots$};
    \node at (6, 1.8) {$w_{l-1}$};
    \node at (7, 1.8) {$w_l$};
    \draw [->] (0, 2.1) -- (0, 2.3);
    \draw [->] (1, 2.1) -- (1, 2.3);
    \draw [->] (2, 2.1) -- (2, 2.3);
    \draw [->] (3, 2.1) -- (3, 2.3);
    \draw [->] (4, 2.1) -- (4, 2.3);
    \draw [->] (6, 2.1) -- (6, 2.3);
    \draw [->] (7, 2.1) -- (7, 2.3);
    \draw (-.3, 2.3) rectangle (7.3, 3.1);
    \node at (3.5, 2.7) {Transformer decoder $\mathcal D$ from GPT-2};
    \draw [->] (0, 3.1) -- (0, 3.3);
    \draw [->] (1, 3.1) -- (1, 3.3);
    \draw [->] (2, 3.1) -- (2, 3.3);
    \draw [->] (3, 3.1) -- (3, 3.3);
    \draw [->] (4, 3.1) -- (4, 3.3);
    \draw [->] (6, 3.1) -- (6, 3.3);
    \draw [->] (7, 3.1) -- (7, 3.3);
    \node at (0, 3.6) {[SOS]};
    \node at (1, 3.6) {$w_1$};
    \node at (2, 3.6) {$w_2$};
    \node at (3, 3.6) {$w_3$};
    \node at (4, 3.6) {$w_4$};
    \node at (5, 3.6) {$\cdots$};
    \node at (6, 3.6) {$w_l$};
    \node at (7, 3.6) {[EOS]};
    
    \node at (9, 0) {$s_1$};
    \node at (10, 0) {$s_2$};
    \node at (11, 0) {$s_3$};
    \node at (13, 0) {$s_5$};
    \node at (14, 0) {$s_6$};
    \node at (15, 0) {$s_7$};
    \draw [->] (9, .3) -- (9, .5);
    \draw [->] (10, .3) -- (10, .5);
    \draw [->] (11, .3) -- (11, .5);
    \draw [->] (13, .3) -- (13, .5);
    \draw [->] (14, .3) -- (14, .5);
    \draw [->] (15, .3) -- (15, .5);
    \draw (8.7, .5) rectangle (9.3, 1.3);
    \draw (9.7, .5) rectangle (10.3, 1.3);
    \draw (10.7, .5) rectangle (11.3, 1.3);
    \draw (12.7, .5) rectangle (13.3, 1.3);
    \draw (13.7, .5) rectangle (14.3, 1.3);
    \draw (14.7, .5) rectangle (15.3, 1.3);
    \node at (9, .9) {$\mathcal E$};
    \node at (10, .9) {$\mathcal E$};
    \node at (11, .9) {$\mathcal E$};
    \node at (13, .9) {$\mathcal E$};
    \node at (14, .9) {$\mathcal E$};
    \node at (15, .9) {$\mathcal E$};
    \draw [->] (9, 1.3) -- (9, 1.5);
    \draw [->] (10, 1.3) -- (10, 1.5);
    \draw [->] (11, 1.3) -- (11, 1.5);
    \draw [->] (13, 1.3) -- (13, 1.5);
    \draw [->] (14, 1.3) -- (14, 1.5);
    \draw [->] (15, 1.3) -- (15, 1.5);
    \node at (9, 1.8) {$f_1$};
    \node at (10, 1.8) {$f_2$};
    \node at (11, 1.8) {$f_3$};
    \node at (12, 1.8) {$\vec{0}$};
    \node at (13, 1.8) {$f_5$};
    \node at (14, 1.8) {$f_6$};
    \node at (15, 1.8) {$f_7$};
    \draw [->] (9, 2.1) -- (9, 2.3);
    \draw [->] (10, 2.1) -- (10, 2.3);
    \draw [->] (11, 2.1) -- (11, 2.3);
    \draw [->] (12, 2.1) -- (12, 2.3);
    \draw [->] (13, 2.1) -- (13, 2.3);
    \draw [->] (14, 2.1) -- (14, 2.3);
    \draw [->] (15, 2.1) -- (15, 2.3);
    \draw (8.7, 2.3) rectangle (15.3, 3.1);
    \node at (12, 2.7) {Sentence-level transformer $\mathcal T$};
    \draw [->] (12, 3.1) -- (12, 3.3);
    \node at (12, 3.6) {$\hat f_4$};
    \end{tikzpicture}
    \caption{Model overview. Left panel: Denoising autoencoder. The encoder $\mathcal E$ takes a corrupted sentence (with each token $w_i$ for $i=1,2,\ldots,l$ masked randomly) as input and outputs a representation of the sentence. The decoder $\mathcal D$ should reconstruct the original uncorrupted sentence. The training parameters of $\mathcal E$ and $\mathcal D$ are initialized with those of BERT and GPT-2 , respectively. Right panel: Sentence-level transformer. Using the encoder $\mathcal E$, we obtain the representation of every contextual sentence. These sentence representations are fed into a sentence-level transformer $\mathcal T$, which outputs a representation of the missing sentence.}
    \label{architecture}
\end{figure*}

\paragraph{Model Overview.} At a high level, our model consists of two components: a (denoising) \textit{autoencoder} and a \textit{sentence-level transformer}. The former maps each sentence to a fixed-length feature vector in the latent semantic space, and reconstructs the sentence from the representation. The latter predicts the semantic features of the missing sentence from those of contextual sentences. We call our model INter-SEntential Transformer (\model).

Formally, let $(\mathcal E, \mathcal D)$ be an autoencoder, where $\mathcal E$ ($\mathcal D$) is the encoder (decoder) such that $\mathcal E\circ \mathcal D$ and $\mathcal D\circ\mathcal E$ are supposed to be identity maps. Let $\mathcal T$ be a sentence-level transformer with positional encoding $\mathcal P$. The transformer $\mathcal T$ takes the contextual information as input and outputs a hypothetical representation of the missing sentence. Specifically,
\begin{align} \label{model}
& \hat  s_m  = \mathcal D\big(\mathcal T(f_1 + \mathcal P(1), f_2 + \mathcal P(2), \ldots,  \nonumber \\
& \, f_{m-1} + \mathcal P(m - 1), \vec{0} + \mathcal P(m), \nonumber \\ 
& \, f_{m+1} + \mathcal P(m + 1), \ldots, f_M + \mathcal P(M))[m]\big), 
\end{align}
where $f_j=\mathcal Es_j$ is the encoding of the sentence $s_j$, $\vec{0}$ is the zero vector representing the missing sentence, and $\mathcal T(\cdots)[m]$ is output of the transformer $\mathcal T$ in the missing position $m$.

The autoencoder and the sentence-level transformer can be trained separately. We first train the former on individual sentences. Then, we precompute and save the feature vectors of all sentences. While training the latter, it is not necessary to load the former. This makes training more efficient.

\paragraph{Sentence Representation Learning via Denoising Autoencoding.}

Large-scale pre-training approaches (e.g., BERT) lead to superior performance in many language understanding tasks related to sentence representation learning \cite{RG19}. However, the features learned by BERT (or fine-tuned on downstream tasks) cannot be directly used for generation tasks, as the masked language model objective of BERT does not enforce the reconstruction of the original sentence from the extracted features. Instead of directly using BERT features, we learn sentence representations via autoencoding. This naturally integrates BERT and GPT-2, and combines sentence representation learning and generation.

As shown in the left panel of Figure \ref{architecture}, we pad the [CLS] token at the beginning of each sentence $s_j$. We initialize the encoder $\mathcal E$ with BERT, and extract the output $f_j$ corresponding to the [CLS] token as the embedding of $s_j$. We initialize the decoder $\mathcal D$ with GPT-2, and feed $f_j$ as the embedding of the zeroth token. Then, we have $\mathcal D$ generate a sequence of tokens in the hope that the sequence matches $s_j$ (padded with special tokens [SOS] at the beginning and [EOS] at the end). To train the autoencoder, we use teacher forcing and minimize the negative log-likelihood loss by (fine-)tuning the parameters of $\mathcal E$ and $\mathcal D$ jointly. 

An autoencoder embeds sentences into vectors in the latent space. We hope that the embedding is smooth in the sense that semantically similar sentences are mapped to vectors that are close to each other. In particular, interpolation between two points in the latent space should correspond to a smooth semantic transition in the text domain. To this end, we use the following two tricks.

First, we employ a denoising autoencoder, which is known to yield a smoother embedding \cite{VLBM08}. To add noise, we randomly mask each token in $s_j$ with probability $15\%$ by replacing the masked tokens with a special token [MASK]. During training, we use the ``noisy'' $s_j$ with masks as input to the encoder, and use the ``clean'' $s_j$ without masks to compute the loss function. Of course, one could try more sophisticated noise-adding strategies \cite{LLG+19}.

Second, we use early stopping. In our experiments, we observe that as training proceeds, the validation loss of the autoencoder keeps decreasing. In the absence of masks, presumably it would eventually decay to zero so that the autoencoder perfectly reconstructs every sentence. However, this does not necessarily imply that the embedding is smooth. On the contrary, an overtrained autoencoder often tries to remember every individual token and thus fails to achieve smoothness in the latent semantic space. Moreover, it can catastrophically forget some of the information in the initial pre-trained model (GPT-2) and partially lose the power of generating fluent sentences. In practice, we select a checkpoint by monitoring its validation performance on sentence interpolation. Some examples of sentence interpolation are shown in Table \ref{tab:interpolation}.

\paragraph{Sentence Feature Prediction.} After encoding sentences into feature vectors, we use a sentence-level transformer $\mathcal T$ to predict the feature vector of the missing sentence from those of contextual sentences. This is analogous to the task of predicting masked tokens for (pre-)training BERT \cite{bert}, but now it is at the sentence level. Indeed, sentence feature vectors in $\mathcal T$ correspond to token embeddings in BERT, and sentence position ID in $\mathcal T$ corresponds to position ID in BERT.

We train the transformer $\mathcal T$ with the objective 
\begin{equation}
\mathcal L_{\rm SentTrans} = 1-\cos(f_m, \mathcal T(\cdots)[m]),
\end{equation}
where $\cos(\cdots)$ is the cosine similarity between the ground truth sentence feature vector $f_m$ and the prediction $\mathcal T(\cdots)[m]$ in Eq. (\ref{model}). Note that $\cos(\cdots)$ is a good similarity measure only when its arguments are unit vectors. This is guaranteed by the technical trick of fixing the parameters of the last LayerNorm of the transformers $\mathcal E$ and $\mathcal T$, i.e., do not compute the gradients of these parameters in backpropagation.

\paragraph{Generating Sentences from Features.}
At test time, we use the decoder $\mathcal D$ to generate the missing sentence by mapping the predicted feature vector to the text domain. Note that standard generation schemes such as top-$k$ sampling, beam search, and nucleus sampling \cite{holtzman2019curious} can be used without additional modeling effort.

\paragraph{Computational Efficiency.} Compared with vanilla GPT-2, our model can process and analyze a document containing many sentences at the discourse level with dramatically lower time and space complexity. To estimate quantitatively, suppose that a document contains $N_s$ sentences, each of which has $N_{t}$ tokens. Then, the time complexity is reduced from $\mathcal{O}(N_s^2N_t^2)$ to $\mathcal{O}(N_s^2+N_sN_t^2)$. Moreover, sentence features can be precomputed once and then reused for every epoch or even in other tasks on the same dataset. If sentence features have been precomputed and are already directly available, the time complexity is further reduced to $\mathcal{O}(N_s^2)$.

\subsection{Sentence Infilling with Lexical Constraints} \label{wkey}

We further introduce a related task called \emph{sentence infilling with lexical constraints}, which is the same as sentence infilling except that now we are given some keywords of the missing sentence as an additional input to hint the generation. The keywords are treated as \emph{soft} constraints (aka priming): The generated sentence is not directly enforced to contain the exact keywords. It may contain a synonym or share some semantics with the keywords.

We expect that the presence of keyword constraints makes the task more difficult rather than easier, although incorporating keywords can significantly improve the BLEU score of the generation with respect to the ground truth. Intuitively, keywords force the model to speculate the semantics of the ground truth sentence, and significantly reduce the number of possible solutions. In the absence of keywords, the model has the freedom of completing the task according to its own way of thinking.

To handle keyword constraints, we introduce a new component called the \emph{constraint feature encoder} to our architecture. It is a transformer encoder $\mathcal K$ that maps a set $S$ of keywords to a feature vector that lives in the same latent space of sentence embeddings. We train $\mathcal K$ with knowledge distillation \cite{kim2016sequence}. The teacher model is the sentence encoder $\mathcal E$, which maps a sentence containing the keywords in $S$ to a feature vector. We use the cosine similarity loss between these two feature vectors to teach the student model $\mathcal K$.

For implementation details, suppose we have two keywords $w_1$ and $w_2$. Then, the input to $\mathcal K$ is three tokens $(\text{[CLS]},w_1,w_2)$. We replace the zero vector in Eq. (\ref{model}), which represents the missing sentence, with the output of $\mathcal K$ above the [CLS] token. We do not use positional encoding in $\mathcal K$ because keywords do not have order.

\section{Experiments}

\subsection{Experimental Setup}

We evaluate our model on two datasets (TripAdvisor and Recipe). We have released the source code to facilitate future research (\url{https://github.com/dreasysnail/INSET}).

\paragraph{Dataset and Preprocessing.} We conduct experiments on the TripAdvisor and Recipe datasets. For the TripAdvisor dataset of hotel reviews \cite{wang2010latent}, we partially follow the preprocessing in \cite{cho-etal-2019-towards}. Our preprocessing includes, but is not limited to: (i) discarding reviews containing non-English tokens; (ii) removing duplicate reviews so that only one copy is retained. We set the maximum number of tokens in a sentence to be $32$ and the minimum number of sentences in a review to be $7$ (so that the context is not too short). Any review with longer sentences or having a smaller number of sentences is discarded.

We use the following strategy to mask sentences. For a paragraph consisting of $M\ge7$ consecutive sentences, we split it into $M-6$ data points, each of which has exactly $7$ sentences. The $j$'th data point spans from the $j$'th to the $(j+6)$'th sentence (inclusive) of the paragraph, for $j=1,2,\ldots,M-6$. We mask the middle (i.e., $4$th) sentence for each data point so that the masking rate is $1/7\approx14.3\%$, which is close to that ($15\%$) of BERT. After preprocessing, the size of the dataset (training, validation, test) is (1108134, 62543, 533) data points.

Our strategy of always masking the middle sentence out of $7$ sentences is not only the simplest but also without loss of generality. Our model is directly applicable to the situation where we randomly mask, e.g., $3$ out of $20$ sentences. However, the quality of human evaluation may be affected because the patience and attention of human evaluators may decrease as the context length increases. For the effectiveness of human evaluation, we use the simplest strategy to mask sentences.

The Recipe dataset is obtained from (\url{https://commoncrawl.org}), where the metadata is formatted according to Schema.org (\url{https://schema.org/Recipe}). We use the same preprocessing as that of the TripAdvisor dataset except that instructions with less than $5$ sentences are discarded. After preprocessing, the size of the dataset (training, validation, test) is (1073886, 56055, 500) data points. Recipe instructions usually describe a 
time-ordered procedure, and thus are ideal for testing the reasoning capability of the model.

\paragraph{Evaluation Metrics.} Following \cite{galley2019grounded, zhang2019dialogpt}, we perform automatic evaluation using standard machine translation metrics, including BLEU \cite{papineni2002bleu}, NIST \cite{doddington2002nist}, and METEOR \cite{lavie2007meteor}. As a variant of BLEU, NIST weights $n$-gram matches by their information gain, and thus penalizes uninformative $n$-grams. We also use Entropy \cite{zhang2018generating} and Dist-$n$ \cite{li2015diversity} to evaluate lexical diversity. See \cite{galley2019grounded} for more details.

BLEU, NIST, and METEOR measure the similarity between the generated sentence and the ground truth. They are not ideal scores for our task because a sentence that is semantically very different from the ground truth could possibly fit the context perfectly. However, it may still be helpful to compute these commonly used scores. It is an important and challenging open problem to design an automatic score that faithfully measures the overall quality of the generation in our task.

\paragraph{Baseline.} Our baseline is the self-attention model for text infilling \cite{ZHX19}. It is a transformer language model with novel positional encoding. The traditional approach of encoding the absolute position of each token is not directly applicable to our task because we do not know in advance the absolute positions of contextual tokens after the missing sentence. To resolve this issue, \cite{ZHX19} divides the text into segments. In the case of only one masked sentence, the first (third) segment consists of contextual tokens before (after) the mask, and the second corresponds to the mask. Then, each token is indexed by its segment ID and its position ID within the segment. The missing tokens are sequentially generated from these IDs and the current surrounding context.

Training the baseline model on our dataset, we use the same set of hyperparameters as in the original reference except that the batch size is set to $250$ (it is $400$ in \cite{ZHX19}). This avoids out-of-memory errors. Note that we are handling much longer sequences (usually $>100$ tokens) than \cite{ZHX19}, in which the maximum number of tokens in a sequence is only $16$.

The baseline model is trained for a sufficient number ($30$) of epochs until the validation (negative log-likelihood) loss and perplexity clearly saturate. We report the results of the checkpoint with the smallest validation loss and perplexity. Note that we observe that other checkpoints in the saturation regime behave very similarly on the test set.

\paragraph{Keyword Extraction.} In the task of sentence infilling with lexical constraints, we need to extract keywords from the masked sentence. Keyword extraction is a classical problem in information retrieval. Standard methods include, but are not limited to, tf-idf (term frequency--inverse document frequency) \cite{ramos2003using}. We have tried tf-idf, but it does not work well for the TripAdvisor dataset of hotel reviews. One reason is that this dataset has quite a few typos, and unfortunately tf-idf favors them because typos occur less frequently than normal words. This issue can be resolved by manually filtering out all typos. After the fix, however, we observe that the quality of extracted keywords remains unsatisfactory.

We use the following strategy to extract keywords. We first define a list of stop words. To this end, we use the stop word list from NLTK \cite{BKL09} and manually add a number of words (e.g., ``hotel'') that are not very informative for the particular dataset of hotel reviews. For each sentence, we select non-stop words that appear most frequently in the entire dataset. We usually select two keywords per sentence, but occasionally select one or even zero if few words remain after filtering out stop words and typos. We observe that the keywords extracted with this strategy can pivot the gist of most sentences well.

\paragraph{Model Size and Hyperparameters.} Our architecture has several components. The encoder $\mathcal E$ and the sentence-level transformer $\mathcal T$ have the same size as BERT$_\text{ BASE}$. The decoder $\mathcal D$ has the same size as GPT-2 (117M). In the presence of lexical constraints, the constraint feature encoder $\mathcal K$ has the same size as BERT$_\text{BASE}$. During decoding, we use beam search with beam size $5$.

\subsection{Experimental Results}

\begin{table}[t!]
\footnotesize
\begin{tabular}{l |p{2.6in} }
\cmidrule[\heavyrulewidth]{1-2}
 & example 1 \\
\cmidrule[\heavyrulewidth]{1-2} 
A & The pool area was nice and sunbathing was great. \\
- &  The pool area was nice and staff was great. \\
- &  The pool area staff was nice and very helpful. \\
- &  Front desk staff were very helpful and friendly. \\
B &  Front desk staff were very nice and helpful. \\
\end{tabular}
\begin{tabular}{l |p{2.6in} }
\cmidrule[\heavyrulewidth]{1-2}
  & example 2 \\
\cmidrule[\heavyrulewidth]{1-2} 
A & The service was attentive and we had the best food in town. \\
- &  The service was attentive and we had a great room with plenty of food. \\
- &  The room was spacious with good service and we had a queen bed. \\
- &  The room was very spacious with queen beds.  \\
B &  The room was very spacious with 2 queen beds.  \\
\bottomrule
\end{tabular}
\caption{Sentence interpolation. ``A'' and ``B'' are two sentences in the test set. The intermediate sentences are generated by interpolating between the latent-space representations of A and B.}
\label{tab:interpolation}
\end{table}

\paragraph{Sentence Representation Learning.} We first qualitatively evaluate the smoothness of the latent-space sentence embeddings learned via denoising autoencoding. Table~\ref{tab:interpolation} shows two examples of sentence interpolation on the TripAdvisor dataset. In each example, the first and last sentences are inputs by hand, and the $3$ intermediate ones are interpolations generated by our (denoising) autoencoder. We observe that the interpolations not only combine words from input sentences, but are readable, meaningful, and show a smooth semantic transition from the first to the last sentence. We speculate that the power of generating fluent and semantically coherent sentence interpolations is derived from BERT and GPT-2. Inherited from these large-scale pre-trained models, the latent-space sentence embedding is reasonably smooth as our sentence interpolation results show.

\begin{table*}[t!]
\centering
\small
\begin{tabular}{p{0.4in}| r H  r H  r | H r  H r | r| H H H r |r  r | r}
\cmidrule[\heavyrulewidth]{1-18}
Dataset &  & \multicolumn{4}{c|}{NIST} & \multicolumn{4}{c|}{BLEU} & MET- & \multicolumn{4}{c|}{Ent.} & \multicolumn{2}{c|}{Dist} & \multicolumn{1}{c} {Len.}  \\
& Method & N-1 & N-2 & N-3 & N-4 & B-1 & B-2 & B-3 & B-4 & EOR & E-1 & E-2 & E-3 & E-4 &  D-1 &D-2 & \\
\cmidrule[\heavyrulewidth]{1-18} 
\multirow{7}{*}{Trip} & \multicolumn{17}{l}{\textit{Without keyword constraints:}}  \\
  & baseline & 0.52 & 0.54 & 0.54 & 0.54 & 16.34\% & 4.29\% & 1.59\% & 0.54\% & 5.85\% & 2.56 & 2.94 & 3.06 & 3.10 & 1.32\% & 2.23\% & 6.97\\
  & \model~(full context) & 1.16 & \textbf{1.23} & 1.23 & \textbf{1.23} & 21.25\% & \textbf{6.08\%} & 2.22\% & \textbf{0.96\%} & \textbf{7.04\%} & 5.37 & 7.17 & 7.92 & \textbf{8.13} & \textbf{16.30\%} & \textbf{46.64\%} & 10.70\\
  & \model~(less context) & 0.98 & 1.02 & 1.02 & 1.02 & 18.77\% & 4.74\% & 1.52\% & 0.51\% & 5.83\% & 5.23 & 6.95 & 7.64 & 7.85 & 12.98\% & 41.39\% & 11.26\\
\hhline{~-----------------}
& \multicolumn{17}{l}{\textit{With keyword constraints:}}\\
& \model~(w/ context) & 2.72 & \textbf{3.09} & 3.14 & \textbf{3.15} & 38.40\% & \textbf{20.14\%} & 11.40\% &\textbf{ 6.57\%} & \textbf{16.48\%} & 5.89 & 7.79 & \textbf{8.30} & \textbf{8.34} & \textbf{22.61\%} & \textbf{63.60\%} & 11.23\\	
& \model~(w/o context) & 2.64 & 3.00 & 3.04 & 3.04 & 37.04\% & 19.47\% & 10.72\% & 6.07\% & 16.00\% & 5.78 & 7.61 & 8.11 & 8.16 & 20.51\% & 57.41\% & 11.12\\
\hhline{~-----------------}
& ground truth (human) & - & - & - & - & - & - & - & - & - & 6.36 & 8.19 & 8.46 & 8.40 & 33.96\% & 79.84\% & 11.36\\
\cmidrule[\heavyrulewidth]{1-18} 
\multirow{3}{*}{Recipe}  & baseline & 0.64 & 0.67 & 0.68 & 0.68 & 13.49\% & 3.91\% & 1.67\% & 0.88\% & 5.23\% & 2.6 & 2.85 & 3.1 & 3.12 & 0.37\% & 0.47\% & 15.32\\
&\model~(ours) & \textbf{1.27} & \textbf{1.36} & \textbf{1.36} & \textbf{1.37} & \textbf{22.14\%} & \textbf{7.24\%} & \textbf{2.88\%} & \textbf{1.33\%} & \textbf{7.07\%} & \textbf{5.57} & \textbf{7.35} & \textbf{7.91} & \textbf{7.99} & \textbf{20.12\%} & \textbf{55.13\%} & 9.63\\
\hhline{~-----------------}
&ground truth (human) & - & - & - & - & - & - & - & - & - & 6.15 & 7.95 & 8.26 & 8.22 & 29.21\% & 74.97\% & 10.55\\
\bottomrule
\end{tabular}
\caption{Automatic evaluation. ``w/ context'' indicates that the generation is based on both keywords and context. ``w/o context'' indicates that the generation is only based on keywords but not context. ``Ent.'' and ``Len.'' stand for Entropy and the average generation length, respectively.}
\label{tab:auto_eval}
\end{table*} 

\begin{table*}[ht!]
    \centering
    \small
    \begin{tabular}{cc|c|ccc}
        \toprule
        system A & system B & criterion & prefer A (\%)& same (\%) & prefer B (\%)\\
        \midrule
        & & coherence & \textbf{54.16} & 13.76 & 32.07 \\
        \model~(ours) & baseline & fluency & \textbf{43.38} & 26.98 & 29.64 \\
        & & informativeness & \textbf{53.48} & 18.79 & 27.72 \\
        \midrule
        & & coherence & 27.87 & 15.69 & \textbf{56.44} \\
        \model~(ours) & ground truth & fluency & 21.78 & 31.38 & \textbf{46.84} \\
        & & informativeness & 27.49 & 21.92 & \textbf{50.59} \\
        \midrule
        \model & & coherence & 18.50 & 23.45 & \textbf{58.04} \\
        w/ keywords & ground truth & fluency & 17.82 & 29.78 & \textbf{52.39} \\
        w/ context & & informativeness & 20.54 & 26.13 & \textbf{53.33} \\
        \midrule
        \model & \model & coherence & \textbf{37.71} & 37.62 & 24.68 \\
        w/ keywords & w/ keywords & fluency & 36.16 & \textbf{37.87} & 25.97 \\
        w/ context & w/o context & informativeness & 35.93 & \textbf{39.86} & 24.21 \\
        \midrule
        \model & \model & coherence & 34.97 & 17.06 & \textbf{47.97} \\
        w/ keywords & w/o keywords & fluency & 29.30 & 28.04 & \textbf{42.65} \\
        w/ context & w/ context & informativeness & 31.73 & 23.24 & \textbf{45.03} \\
        \bottomrule
    \end{tabular}
    \caption{Human evaluation. ``w/(w/o) keywords'' and ``w/(w/o) context'' indicate whether the generation is based on keywords and context, respectively. All numbers are percentages.} 
    \label{tab:human_eval}
\end{table*}

\paragraph{Automatic Evaluation.} Table \ref{tab:auto_eval} shows the BLEU, NIST, METEOR, Entropy, Dist-$n$ scores, and the average length (number of words) of the generated sentences. For the TripAdvisor dataset, we also present results in the presence of keyword constrains.

Table~\ref{tab:auto_eval} compares the baseline \cite{ZHX19}, our results, and the ground truth. In the absence of keyword constraints, \model~outperforms the baseline in terms of all scores on both datasets. This indicates that our results are semantically closer to the ground truth and are more diverse than the baseline. In terms of the average generation length, our results are much closer to the ground truth than the baseline is.

Table~\ref{tab:auto_eval} also presents two ablation studies. The first shows the performance decrease with less context. Recall that each data point in the TripAdvisor dataset has $6$ contextual sentences (full context). We train \model~on the same set of data points but truncate the context to $4$ sentences (less context). The second ablation study shows the effect of context in the presence of keywords. We compare two models. The first (\model~w/ context) is the model described in Subsection \ref{wkey}. Its generation is based on both keywords and context. The second model (\model~w/o context) is $\mathcal D\circ\mathcal K$, which directly decodes the output of the constraint feature encoder $\mathcal K$ using the decoder $\mathcal D$. Its generation is only based on keywords but not context. We observe that the scores of the first model are higher than those of the second. Both ablation studies show that our model can make full use of context to improve the generation.

\begin{table*}[ht!]
    \footnotesize
    \begin{tabular}{p{.1\linewidth}|p{.27\linewidth}|p{.27\linewidth}|p{.25\linewidth}}
        \toprule
        & example from TripAdvisor dataset & example from TripAdvisor dataset & example from Recipe dataset \\
        \midrule
        preceding context & It was such a pleasure to see somthing new every night. It was not very crowded so we were able to get great seats at either the pool or the beach. The VIP sevice was great for dinner reservations and pillow service. & The walls are very thin. Since this is a family vacation type of hotel, people are up at the pool/bbq area/hallways during all hours of the night. Do not stay here if you need a quite night of sleep.  &  After another 15 minutes or so the mixture should thicken up. The mixture will continue to thicken as it cools. \\
        \midrule
        following context & Enjoyed the shrimp coctail and seafood salad delivered to us while enjoying the pool. All of us would not want to stay at another resort and are planning to go back again. Enjoy and Hola!Karen and FriendsMilford, CT & You have to take multiple elevators to go all the way to the 5th floor. My other complaint is that the hotel staff seemed a bit unprofessional. Not what I'm used to when I stay at Marriot properties. & Sterilize your jars and lids and while still hot fill with the jam leaving about a 1/2 inch headspace. Place lids onto the jars and boil in a water bath with jars covered by 3 inches of water for 10 minutes.\\
        \midrule
        ground truth & We did bring a lot of \$1 for tipping and of course the service stepped up a notch more. & Also, the elevator situation is weird. & Remove from the heat and stir in your amaretto.\\
        \midrule
        baseline & The staff was friendly and helpful. & The rooms are very clean and well kept. & Add the flour mixture to the dry ingredients and mix well. \\
        \midrule
        \model  & The buffet dinner was amazing and we had the best food in the resort. & There is only one elevator block in the hotel. & Carefully remove the jars from hot water and keep going until a thick sauce is formed. \\
        \midrule \midrule
        + keywords & \$, service & elevator, situation & - \\
        \midrule
        \model~(w/ keywords) & Service fee for the buffet dinner was \$5.00 and we paid \$5.00 extra for food service. & The elevator situation is extremely frustrating. & -\\
        \bottomrule
    \end{tabular}
    \caption{Examples generated by our model and the baseline.}
    \label{generated_examples}
\end{table*}

\begin{table*}[ht!]
    \footnotesize
    \begin{tabular}{p{.15\linewidth}|p{.79\linewidth}}
        \toprule
        preceding context & My room was a very good size. Tiled floors and woodchip painted walls. The tv did not work - so what.\\
        \midrule
        following context & Great places to eat close by and very reasonable. No air con -so summer could be sticky. My concern is the left luggage room not supervised.\\
        \midrule
        human oracle & The location is terrific beside Sevilla metro stn so only  2 to get by metro all the way to airport. \\
        \midrule 
        + (walk, shopping) &  Walking distance to shopping mall and Circular Quay. \\
        \midrule
        + (internet, \$) &  Internet cost \$20.00 per day. \\
        \bottomrule
    \end{tabular}
    \caption{Examples generated by our model in the same context but with different keywords. ``+ ($\cdots$)'' is keywords.}
    \label{tab:generated_keys}
\end{table*}

\paragraph{Human Evaluation.} We performed human evaluation of our method on the TripAdvisor dataset. We used a crowd evaluation platform to compare two systems and assess their fluency, informativeness, and relevance to the surrounding context (coherence) on $500$ random samples from the test set. Following recommended best practices, each sample was evaluated by five judges. We performed simple spam detection by excluding judges that were too fast or performed too low on a gold set. To avoid bias, we randomized the position of each system while asking judges to compare our systems (with and without keywords) with the ground truth and the text infilling baseline \cite{ZHX19}.

Table \ref{tab:human_eval} shows the human evaluation results. The judges strongly prefer our results (without keywords) to the baseline in all aspects: coherence, fluency, and informativeness. They also strongly prefer the ground truth to our results. Moreover, our results with keywords and context are compared with three other systems: (i) the ground truth; (ii) our results with keywords but not context; (iii) our results with context but not keywords. The second comparison shows that in the presence of keywords, our model can use context to improve all aspects of the generation. The third comparison shows that the presence of keywords reduces the performance of our model, probably because keywords are constraints that the model must take care of.

\paragraph{Generated Examples.} To qualitatively demonstrate the effectiveness of our model, Table~\ref{generated_examples} shows some examples from the TripAdvisor and Recipe datasets. We observe that the baseline \cite{ZHX19} tends to generate generic sentences, while our results (either with or without keywords) are more informative and can fit the surrounding context reasonably well. Table~\ref{tab:generated_keys} shows examples generated by our model in the same context but with different keywords. Our model can extend keywords to a full sentence, adapting to the context. More examples generated by our model on both datasets are given in Appendix \ref{app}.

\section{Conclusions and Outlook}

We study the task of sentence infilling, which is analogous to the masked language modeling task for (pre-)training BERT, but now it is at the sentence level. Sentence infilling requires the model to handle long-range inter-sentential correlation and to process high-level semantic information. It is complementary to (token-level) masked language modeling, which focuses more on syntactic appropriateness and short-range correlation. We propose a framework called \model~to decouple three aspects of the task (understanding, planning, and generation) and address them in a unified manner. We demonstrate the effectiveness of our approach using automatic and human evaluation.

Our approach can be modified or extended in some ways. (i) We use a denoising autoencoder to obtain sentence embeddings. One can try to use a variational autoencoder \cite{KW14} instead. A large-scale pre-trained variational autoencoder \cite{li2020optimus} could possibly improve the smoothness of sentence embeddings. (ii) Our model predicts a feature vector for the missing sentence. This vector can be fed into and serve as a guide to token-level models including the baseline \cite{ZHX19}.

Since sentence infilling is analogous to masked language modeling, we expect that it can also be used as a pre-training task. For example, in machine translation of long texts, it is often the case that sentences are translated independently from each other. This sometimes leads to incoherence or even inconsistency between the translated sentences. A post-editor to fix the issue \cite{VST19} should be able to understand inter-sentential relationship and to generate fluent sentences in the 
surrounding context, both of which can be learned from sentence infilling.

\paragraph{Note.} After this paper was posted on arXiv, some related works appeared. \cite{SQBJ20} proposes Blank Language Model for text infilling and other tasks. \cite{DLL20} trains (fine-tunes) a language model (GPT-2) for text and sentence infilling. \cite{li2020optimus} pre-trains a large-scale variational autoencoder with a pair of BERT and GPT-2. \cite{IGEC20} uses a sentence-level language model, which operates on sentence embeddings obtained from BERT, to predict story endings.

\section*{Acknowledgments}

We thank Bill Dolan, Chris Quirk, and Jingjing Liu for helpful discussions and suggestions.

\bibliography{sentfill}
\bibliographystyle{acl_natbib}

\appendix

\section{Additional Generated Examples} \label{app}

Tables \ref{tab:trip_examples}, \ref{tab:recipe_examples} show some additional examples generated by our model (without keywords) on the TripAdvisor and Recipe datasets, respectively. The results are semantically informative and can fit the surrounding context reasonably well. Table \ref{tab:generated_keys_supp} provides additional examples to Table \ref{tab:generated_keys}. Our model can incorporate keywords into the generated sentence in a smart way, adapting to the context.

\begin{table*}
    \footnotesize
    \begin{tabular}{p{.1\linewidth}|p{.41\linewidth}|p{.41\linewidth}}
        \toprule
        & example 1 & example 2 \\
        \midrule
        preceding context & I went in October to meet with their FABULOUS wedding coordinator Summer Laetari. Their property is very beautiful, it's extremely green and lush. Parrot Key has 4 pools. & Good Location if traveling for business or you have a car! Got this hotel thru a discount travel company and paid \$65.00 american a night. Excellent deal at this price.\\
        \midrule
        following context & Their cottages are brand new, very clean and well appointed. If you are looking for a place to have a destination wedding I would recommend Parrot Key! My family and I have already planned another trip to visit next month. & Unfortunetly the view is going to be partly blocked with yet another ``Glass tower'' going in. The room was spacious and clean. No tub in our room.\\
        \midrule
        ground truth & It's very colorful and unique. & We had a terrific view from the 16th floor.\\
        \midrule
        \model  & There is also a beach resort with lots of loungers. & We had a room on the upper floor which overlooks the lobby. \\
        \bottomrule
    \end{tabular}
    \begin{tabular}{p{.1\linewidth}|p{.41\linewidth}|p{.41\linewidth}}
        \toprule
        & example 3 & example 4 \\
        \midrule
        preceding context & My family stayed here for 5 nights in August 2011. The resort is beautiful and the grounds are immaculately manicured. The kitchen is great for the family. & We stayed in 2 interconnecting rooms as we are a family of 5. We started off with a bad start, as the check in was not aware that we were with 3 kids. I booked directly with them and got a confirmation via email for 2 rooms for 2 adults.\\
        \midrule
        following context & We would just pack a cooler and head out in our rental car and explore the island. The pools at the resort were fabulous and the staff was attentive. We used the grills(kept very clean) several nights. & Obviously this was not reflected in the paper work check-in had. We could only add an extra bed for an extra charge, but I refused to pay for this as I had phoned them before. The check-in lady would not bend, and we had to go for 2 rooms with 2 seperate beds.\\
        \midrule
        ground truth & We were able to keep essentials in the room which made those early morning excursions more enjoyable. & Before we arrived I called reservations to change this into 2 adults and 3 children.\\
        \midrule
        \model  & We have plenty of kitchen utensils and the beach was a nice place to stay. & When we checked in we were told that we had to request another room on the 2nd floor due to the extra charges. \\
        \bottomrule
    \end{tabular}
    \begin{tabular}{p{.1\linewidth}|p{.41\linewidth}|p{.41\linewidth}}
        \toprule
        & example 5 & example 6 \\
        \midrule
        preceding context & It was such a pleasure to see somthing new every night.  It was not very crowded so we were able to get great seats at either the pool or the beach.  The VIP sevice was great for dinner reservations and pillow service. & My intentions were to expect the worst which made my stay there that much better than everyone elses. If everyone thought they were staying at the Hyatt, no wonder they thought so negatively about the place. I am in my late twenties and wanted a place where I could walk to local bars, restaurants, etc. \\
        \midrule
        following context & Enjoyed  the  shrimp  coctail  and  seafood salad delivered to us while enjoying the pool. All of us would not want to stay at another resort and are planning to go back again. Enjoy and Hola!Karen and FriendsMilford, CT & This was the perfect place for me. As far as the accomodations, the beds were small (but so was everywhere else in Europe) and the showers were unusual. Otherwise it was worth the money for a prime time location in the heart of the night life area. \\
        \midrule
        ground truth & We did bring a lot of \$1 for tipping and of course the service stepped up a notch more. & without struggling to find my way home at night. \\
        \midrule
        \model & The buffet dinner was amazing and we had the best food in the resort. & So I had no reason to stay in the HOTEL itself. \\
        \bottomrule
    \end{tabular}
    \caption{Generated examples by our model on the TripAdvisor dataset}
    \label{tab:trip_examples}
\end{table*}

\begin{table*}
    \footnotesize
    \begin{tabular}{p{.1\linewidth}|p{.41\linewidth}|p{.41\linewidth}}
        \toprule
        & example 1 & example 2 \\
        \midrule
        preceding context & Roll up rectangles width-wise and pinch ends to seal. Bake for 12 minutes or until the tops begin to brown. & Drizzle each potato cup with 1 teaspoon browned butter. Cover muffin tin tightly with aluminium foil and place in oven.\\
        \midrule
        following context & Best when served warm. For added flavor, serve with strawberry jelly. & Remove from oven and turn broiler on high. Sprinkle potato rounds evenly with remaining parmesan cheese.\\
        \midrule
        ground truth & Let cool on baking sheet. & Bake for 25 minutes.\\
        \midrule
        \model & Cool on wire rack and remove. & Bake for 20 minutes or until potatoes are tender. \\
        \bottomrule
    \end{tabular}
    \begin{tabular}{p{.1\linewidth}|p{.41\linewidth}|p{.41\linewidth}}
        \toprule
        & example 3 & example 4 \\
        \midrule
        preceding context & Preheat oven to 425 degrees Fahrenheit. Line a baking sheet with a SILPAT mat. & Heat the oil in a pan at medium. Add the mushrooms and saute until tender, about 7-10 minutes.\\
        \midrule
        following context & With a pastry cutter, cut in the coconut oil and the butter. Make a well and add in the milk 1/2 cup at a time, stirring gently with a wooden spoon. & Add the reserved water and simmer at medium-high until reduced by half, about 10 minutes. Meanwhile cook the pasta as directed on the package.\\
        \midrule
        ground truth & In a bowl, mix the flour, baking powder, baking soda and sea salt. & Add shallots, garlic, thyme, salt and pepper and saute for 2 minutes.\\
        \midrule
        \model  & In a medium bowl, mix together the flour, baking powder, sugar, salt and cinnamon. & Add the garlic and sautee until fragrant, about 2 minutes. \\
        \bottomrule
    \end{tabular}
    \begin{tabular}{p{.1\linewidth}|p{.41\linewidth}|p{.41\linewidth}}
        \toprule
        & example 5 & example 6 \\
        \midrule
        preceding context & After another 15 minutes or so the mixture should thicken up. The mixture will continue to thicken as it cools. & Bake the graham cracker crust for 10 minutes. Remove from oven and allow to cool to room temperature.\\
        \midrule
        following context & Sterilize your jars and lids and while still hot fill with the jam leaving about a 1/2 inch headspace. Place lids onto the jars and boil in a water bath with jars covered by 3 inches of water for 10 minutes. & Stir in the lime zest and lime juice. Stir until mixture is smooth and begins to slightly thicken.\\
        \midrule
        ground truth & Remove from the heat and stir in your amaretto. & Meanwhile, combine the egg yolks and condensed milk in a medium bowl.\\
        \midrule
        \model  & Carefully remove the jars from hot water and keep going until a thick sauce is formed. & In a medium bowl, combine the cream cheese and powdered sugar, stirring until smooth. \\
        \bottomrule
    \end{tabular}
    \caption{Generated examples by our model on the Recipe dataset}
    \label{tab:recipe_examples}
\end{table*}

\begin{table*}
    \footnotesize
    \begin{tabular}{p{.16\linewidth}|p{.78\linewidth}}
        \toprule
        preceding context & Also has a safe. The hotel is in a good location, beside the City Centre and there are a nice selection of shops within the Monte Carlo. Service was very good but avoid the concierge in the morning when people are booking tours, the queues are long.\\
        \midrule
        following context & No wi-fi in the room which is a bit annoying but they have it in the foodcourt by Starbucks and McDs. Also we were disappointed to see the \$15/night resort fee was charged to our credit card after our stay. I don't recall them mentioning this at check-in. \\
        \midrule 
        human oracle & CVs is next door and it's 24/7 so you can buy snacks and anything else you fancy. \\
        \midrule 
        + (breakfast, cereal) & Breakfast is included with cereal, muffins and breads. \\
        \midrule
        + (food, expensive) & Prices are expensive but food in the hotel is very cheap. \\
        \bottomrule
    \end{tabular}
    \caption{Examples generated by our model in the same context but with different keywords. ``+ ($\cdots$)'' is keywords.}
    \label{tab:generated_keys_supp}
\end{table*}

\end{document}